\documentclass{llncs}
\usepackage{llncsdoc}

\usepackage[center,justification=justified,singlelinecheck=true,indention=1em,margin=10pt]{caption}

\renewcommand{\phi}{\varphi}
\usepackage{color,graphicx}
\usepackage{amssymb}
\usepackage{amsmath}
\usepackage{array}
\usepackage{mathtools}
\usepackage{tabularx}
\mathtoolsset{showonlyrefs}
\newcounter{rowcount}
\title{Automated Reasoning for Robot Ethics\thanks{This is a short version of \cite{DBLP:conf/miwai/FurbachSS14}. It is supported by the DFG grants FU~263/15-1 and STO~421/5-1 'Ratiolog'.}}
\author{Ulrich Furbach\inst{1}, Claudia Schon\inst{1} and Frieder Stolzenburg\inst{2}}
\institute{
	Universit\"at Koblenz-Landau, 
	\email{\{uli,schon\}@uni-koblenz.de}
\and
	Harz University of Applied Sciences,
	\email{fstolzenburg@hs-harz.de}
}

\begin{document}
\maketitle

\begin{abstract}
Deontic logic is a very well researched branch of mathematical logic and
philosophy. Various kinds of deontic logics are considered for different
application domains like argumentation theory, legal reasoning, and acts in
multi-agent systems. In this paper, we show how standard deontic logic can be used to model ethical codes for multi-agent systems. Furthermore we show how Hyper, a high performance theorem prover, can be used to prove properties of these ethical codes.
\end{abstract}

\section{Introduction} 

Deontic logic is a very well researched branch of mathematical logic and•
philosophy. Various kinds of deontic logics are considered for different
application domains like argumentation theory, legal reasoning, and acts in
multi-agent systems \cite{handbook}. 

%
This paper applies automated reasoning in standard deontic logic
(SDL) to a problem in multi-agent systems. For this, we follow the example from \cite{journals/expert/BringsjordAB06} where reasoning with ethical codes is presented.
Instead of
implementing a reasoning system for deontic logic directly, we  rely on the first order reasoning system Hyper \cite{WernhardPelzer}, which uses the
hypertableau calculus from 
\cite{DBLP:conf/jelia/BaumgartnerFN96}. This has the advantage that reasoning in deontic logic can be embedded easily into other applications for which Hyper is already an inference system (e.g. the LogAnswer system \cite{DBLP:journals/aicom/FurbachGP10}).

SDL corresponds to the modal logic $\mathsf{K}$ with a seriality axiom. Due to the fact that modal logic $\mathsf{K}$ is (more or less) a notational variant of description logic  $\mathcal{ALC}$,  deontic logic can be translated into it \cite{DBLP:conf/miwai/FurbachSS14}.  Since Hyper offers a decision procedure for various description logics, it can be employed for reasoning in SDL.


In Section \ref{sec:modal} we shortly depict the reasoning framework, and in Section~\ref{sec:multi-agent} we
show how  to model our small example.  This hopefully documents the
applicability of reasoning with SDL in multi-agent systems.
  
\section{Automated Reasoning for Standard Deontic Logic}\label{sec:modal}

Standard deontic logic (SDL) is obtained from the well-known modal logic
$\mathsf{K}$ by adding the seriality axiom $\mathsf{D}$: $\Box P \rightarrow \Diamond P$.
In this logic, the $\Box$-operator is interpreted  as `it is obligatory that'
and the $\Diamond$ as `it is permitted that'. The $\Diamond$-operator can be
defined by $\Diamond P \equiv \neg\Box\neg P$.
The seriality axiom in SDL states that, if a formula  has to hold in all reachable
worlds, then there exists such a world. With the deontic reading of $\Box$ and
$\Diamond$ this means: Whenever the formula $P$ ought to be, then there
exists a world where it holds. In consequence, there is always a world, which is
ideal in the sense that all the norms formulated by `the ought to
be'-operator hold.

Deontic logic is the logic of choice when formalizing knowledge about norms like  
the representation of legal knowledge or ethical codes for agents. However, there are only few automated
theorem provers specially dedicated for deontic logic and used by deontic logicians (see
\cite{Artosi94ked:a,Bassiliades:2011:MDR:2441484.2441486}). Nonetheless,
numerous approaches to translate modal logics into (decidable fragments
of) first-order predicate logics are stated in the literature. A nice overview
including many relevant references is given in \cite{SH13}. 

In this paper, we  use the first order predicate logic theorem prover  Hyper
\cite{WernhardPelzer} to handle SDL knowledge bases. By using the well-known
translation of modal logic into description logic
\cite{Schild91acorrespondence} (called $\phi$ in the sequel) and the fact that 
Hyper offers a decision procedure for the description logic $\mathcal{SHIQ}$ \cite{cadesd}, we are able to process SDL efficiently. In the following, SDL is translated into  $\mathcal{ALC}$, which is a subset of $\mathcal{SHIQ}$.

\subsection*{Transformation from SDL into $\mathcal{ALC}$}
For a normative system consisting of the set of deontic logic formulae
$\mathcal{N}= \lbrace F_{1},\ldots, F_{n}\rbrace$,
the translation $\phi$ is defined as the conjunctive combination of the translation of all deontic logic formulae in $\mathcal{N}$:
\begin{equation}
\phi(\mathcal{N}) = \phi(F_{1}) \sqcap \ldots \sqcap \phi(F_{n})
\end{equation}

Note that $\phi(\mathcal{N})$ does not yet contain the translation of the seriality axiom. As shown in \cite{Klarman28052013}, the seriality axiom can be translated into the TBox  
\begin{equation*}
\mathcal{T}= \lbrace \top \sqsubseteq \exists r.\top\rbrace
\end{equation*}
with $r$ the atomic role introduced by the mapping $\phi$.

For our application, the result of the translation of a normative system $\mathcal{N}$ and the seriality axiom is an $\mathcal{ALC}$ knowledge base $\Phi(\mathcal{N})=(\mathcal{T},\mathcal{A})$,
where the TBox $\mathcal{T}$ consists of the translation of the seriality axiom and the ABox
$\mathcal{A}= \lbrace (\phi(\mathcal{N}))(a)\rbrace$ for a new individual $a$. In description logics, performing a satisfiability test of a concept $C$ w.r.t. a TBox  is usually done by adding a new individual $a$ together with the ABox assertion $C(a)$. For the sake of simplicity, we do this construction already during the transformation of $\Phi$ by adding $(\phi(\mathcal{N}))(a)$ to the ABox.

An advantage of the translation of deontic logic formulae into an
$\mathcal{ALC}$ knowledge base is the existence of a TBox in $\mathcal{ALC}$. 
This makes it possible to add further axioms to the TBox. For example we can add
certain norms that we want to be satisfied in \emph{all} reachable worlds into the
TBox.

\subsection*{Reasoning Tasks}\label{sec:reasoning tasks}
With the help of Hyper, we can solve the following  interesting reasoning tasks:
%
\paragraph{Consistency checking of normative systems:} In practice, normative systems can be very large. Therefore it is not easy to see, if a given normative system is consistent. The Hyper theorem prover can be used to check consistency of a normative system $\mathcal{N}$. 
\paragraph{Evaluation of normative systems:} Given several normative systems, we use Hyper to find out for which normative system a desired outcome is guaranteed. 

\section{An Example from Multi-agent Systems}
\label{sec:multi-agent}
In multi-agent systems, there is a challenging  area of research, namely the formalization of `robot ethics'. It aims at defining formal rules for  the behavior of  agents and to prove certain properties. As an  example  consider Asimov's laws, which aim at regulating the relation between robots and humans. In \cite{journals/expert/BringsjordAB06}, the authors depict  a small example of two surgery robots obeying ethical codes concerning their work. These codes are expressed by means of MADL, which is an extension of standard deontic logic with two operators.
In \cite{Murakami04utilitariandeontic}, an axiomatization of MADL is given. Further it is asserted, that MADL is not essentially different from standard deontic logic. This is why we use SDL to model the example.

\subsection*{Formalization in SDL}\label{sec:formalizationindeonticlogic}
In our example, there are two robots $\mathit{ag1}$ and $\mathit{ag2}$ in a
hospital. For the sake of simplicity, each robot can perform one specific action:
$\mathit{ag1}$ can terminate a person's life support and $\mathit{ag2}$ can
delay the delivery of pain medication. In \cite{journals/expert/BringsjordAB06},
four different ethical codes $\mathit{J}$, $\mathit{J}^{\star}$, $\mathit{O}$
and $\mathit{O^{\star}}$ are considered:
\begin{itemize}
\item ``If ethical code $\mathit{J}$ holds, then robot $\mathit{ag1}$ ought to take care that life support is terminated.'' This is formalized as:
\begin{equation}
\mathit{J}\rightarrow \Box  \mathit{act(ag1, term)}
\end{equation}
\item ``If ethical code $\mathit{J^{\star}}$ holds, then code $\mathit{J}$ holds, and robot $\mathit{ag2}$ ought to take care that the delivery of pain medication is delayed.'' This is formalized as:
\begin{equation}
 \mathit{J^{\star}}\rightarrow \mathit{J} \land \mathit{J^{\star}} \rightarrow \Box  \mathit{act(ag2, delay)}
\end{equation}
\item ``If ethical code $\mathit{O}$ holds, then robot $\mathit{ag2}$ ought to take care that delivery of pain medication is not delayed.'' This is formalized as:
\begin{equation}
\mathit{O}\rightarrow \Box \lnot  \mathit{act(ag2, delay)}
\end{equation}
\item ``If ethical code $\mathit{O^{\star}}$ holds, then code $\mathit{O}$ holds, and robot $\mathit{ag1}$ ought to take care that life support is not terminated.'' This is formalized as:\\
\begin{equation}
\mathit{O^{\star}}\rightarrow \mathit{O} \land \mathit{O^{\star}} \rightarrow \Box \lnot  \mathit{act(ag1, term)}
\end{equation}
\end{itemize}

Further we give a slightly modified version of the evaluation of the acts of the robots, as stated in \cite{journals/expert/BringsjordAB06}, where $(+!!)$ denotes the most and $(-!!)$ the least desired outcome. Note that terms like $(+!!)$ are just propositional atomic formulae here.
\mathtoolsset{showonlyrefs=false}
\begin{align}
\mathit{act(ag1, term)} \land \phantom{\lnot} \mathit{act(ag2, delay)} & \rightarrow (-!!)\label{equ:eval1}\\
\mathit{act(ag1, term)} \land \lnot \mathit{act(ag2, delay)} & \rightarrow (-!)\label{equ:eval2}\\
\lnot \mathit{act(ag1, term)} \land \phantom{\lnot} \mathit{act(ag2, delay)} & \rightarrow (-)\label{equ:eval3}\\
\lnot \mathit{act(ag1, term)} \land \lnot \mathit{act(ag2, delay)} & \rightarrow (+!!)\label{equ:eval4}
\end{align}
\mathtoolsset{showonlyrefs=true}

These formulae evaluate the outcome of the robots' actions. It makes sense to assume, that this evaluation is effective in all reachable worlds. This is why we add formulae stating that formulae~\eqref{equ:eval1}--\eqref{equ:eval4} hold in all reachable worlds. For example, for \eqref{equ:eval1} we add:
\begin{equation}
\Box (\mathit{act(ag1, term)} \land \mathit{act(ag2, delay)} \rightarrow (-!!))\label{equ:allworlds}
\end{equation}
Since our example does not include nested modal operators, the formulae of the form \eqref{equ:allworlds} are sufficient to spread the evaluation formulae to all reachable worlds.
The normative system $\mathcal{N}$ formalizing this example consists of the formalization of the four ethical codes and the formulae for the evaluation of the robots actions.

\subsection*{Reduction to a Satisfiability Test}
A possible query would be to ask if the most desirable outcome $(+!!)$ will come to pass if ethical code $O^{\star}$ is operative. This query can be translated into a satisfiability test. If 
$\mathcal{N} \land \mathit{O^{\star}} \land \Diamond \lnot (+!!)$
 is unsatisfiable, then ethical code $O^{\star}$ ensures outcome $(+!!)$.

Hyper can be used for this satisfiability test. We obtain the desired result namely that (only) ethical code $O^\star$ leads to the most desirable behavior ($+!!$).

\subsection*{Experiment}
We formalized this  example and tested it with the Hyper theorem prover as described above. Since all formalizations are available in $\mathcal{ALC}$, we used the description logic reasoner Pellet \cite{pellet-websem5} to compare the performance with Hyper. It took Pellet 2.548 seconds and Hyper 2.596 seconds to show the unsatisfiability.

Additional experiments can be found in \cite{DBLP:conf/miwai/FurbachSS14}. 
For the examples we considered, the runtimes of Pellet and Hyper are
comparable. 
Further investigation and comparison with other modal and/or description logic reasoning tools is required and subject of future work.

\section{Conclusion}
In this paper, we illustrated with the help of an example how to use standard deontic logic to model ethical codes for multi-agent systems. For normative systems described with deontic logic, there is a translation into description logic concepts. These concepts can be checked automatically by automated theorem provers. We used the Hyper theorem prover to prove that a specified desired outcome is guaranteed, if a certain ethical code is operative. 

\bibliographystyle{abbrv}
\bibliography{wissen}
\end{document}